\documentclass[runningheads]{llncs}
\usepackage[T1]{fontenc}
\usepackage{graphicx}
\usepackage{hyperref}
\usepackage{booktabs}
\usepackage{caption}
\usepackage{multirow}
\usepackage{makecell}
\usepackage{subcaption}

\usepackage{xcolor}
\newcommand{\alg}[1]{\textsc{#1}}
\usepackage{subcaption}
\usepackage{amsmath, amssymb}
\usepackage{booktabs}
\usepackage{todonotes}
\usepackage{tikz}
\usepackage{tikzsymbols}
\usetikzlibrary{arrows,automata}
\usetikzlibrary{shapes.geometric, arrows.meta}
\usetikzlibrary{positioning}
\usetikzlibrary{calc}

\begin{document}
\title{Optimizing Sparse Boolean Rule Networks using Mixed Integer Linear Programming}
\titlerunning{Optimizing Sparse Rule Networks using MILP}
%
%
\author{Thomas Pammer\inst{1}\orcidID{0009-0009-1140-1890} \and
Johannes Fürnkranz \inst{1,2}\orcidID{0000-0002-1207-0159} \and
Markus Sinnl\inst{3}\orcidID{0000-0003-1439-8702}}
\authorrunning{Pammer et al.}
%
\institute{{Institute for Application-Oriented Knowledge Processing (FAW)} 
\and
LIT AI Lab\\
\and
Institute of Business Analytics \& Technology Transformation/JKU Business School\\ Johannes Kepler University Linz, Austria}
\maketitle              
\begin{abstract}
We study the problem of learning accurate and cognitively simple binary classifiers that are directly interpretable as Boolean rules.
We propose mixed integer linear programming (MILP) formulations for training sparse disjunctive normal form (DNF) networks, viewed as small networks
with a conjunctive hidden layer and a disjunctive output layer.
Contrary to previous work that uses MILP for inductive rule learning, we start with a fixed network structure, for which we then find optimal weights.
A regularized objective explicitly trades off misclassification costs against sparsity, yielding globally optimized rule sets.
Our main technical contribution is an enhanced formulation that substantially reduces the number of integer decision variables and eliminates redundant constraints, while preserving exact Boolean semantics for all feasible solutions. Furthermore, we develop a combined approach based on initialization and dropout to break symmetries in the weights between rules, mitigating redundancy, and encouraging more diverse and stable rule representations.
Experiments on synthetic benchmarks and UCI datasets show that the advanced formulation improves computational efficiency and achieves competitive or superior accuracy
compared to established interpretable baselines, including SAT-based DNF learning and well-known  rule- and tree-based learners. 
Overall, our approach provides a principled optimization framework for learning compact Boolean rule sets with explicit control over accuracy and sparsity trade-offs.

\end{abstract}

\section{Introduction}
The importance of interpretable classifiers is gaining attention in machine learning, with strong links to logic, optimization, and knowledge representation.
Although neural networks achieve remarkable predictive performance, their black-box nature often limits their practical applicability in domains where transparency is essential. A promising direction toward interpretable models is to represent classifiers as Boolean circuits, where outputs arise from logical combinations of input features.
Such circuits 
are structurally simple, semantically clear, and align naturally with rule-based reasoning.
Among various representations, \emph{disjunctive normal form} (DNF) structures, which essentially represent sets of rules, are particularly appealing for their interpretability and compatibility with efficient reasoning.
However, exact learning of Boolean circuits from data remains difficult because the space of possible configurations grows exponentially with the number of features and constraints.

\textit{Mixed integer linear programming} (MILP) provides a principled way to combine logical consistency with optimization.
By embedding Boolean relationships directly into linear constraints, MILP enables the learning of accurate and compact classifiers with explicit control of sparsity and model complexity.
Our approach employs a parallel disjunctive structure where multiple conjunctive nodes act simultaneously, and the classifier predicts a positive outcome whenever any learned rule is satisfied.

We introduce two complementary MILP formulations for learning such Boolean circuit classifiers and we additionally compare two slightly different variants of the more advanced formulation.
They can be viewed as small rule networks, where input features feed into hidden conjunctive nodes connected by a disjunction to the output.
Each hidden node defines a rule of the form ``if the active features are present, then predict positive'', where the activation of a feature depends on its weight.
In the basic formulation, hidden nodes are restricted to take binary values, whereas in the advanced formulations they are either relaxed to continuous variables or handled in aggregated constraints.
Using reformulated constraints, parts of the decision variables can be removed, further improving the computational efficiency of both advanced formulations.
Empirical results on synthetic and benchmark datasets show that we achieve competitive accuracy.

Our contributions are summarized as follows:
\begin{itemize}
\item A basic MILP formulation for learning interpretable classification rules. 
\item An advanced formulation that improves efficiency while maintaining interpretability.
\item A reformulation of the advanced MILP formulation using aggregated constraints to improve computational efficiency.
\item A comparison with rule-, decision tree- and SAT-based methods demonstrating competitive accuracy and exact optimization.
\item A combined approach using informed initialization and variable reduction for our advanced formulation to tackle larger problems.
\end{itemize}
The remainder of this paper is structured as follows. Section~\ref{sec:related} shows an overview of related work. Section~\ref{sec:sparse_bool_rn} formally introduces sparse Boolean rule networks, while Section~\ref{sec:milp_formulation} presents the MILP formulations developed in this paper. Section~\ref{sec:exp_eval} reports the computational results obtained with the proposed MILP models and compares them with well-established classification algorithms. Section~\ref{sec:matheuristic} introduces a novel combined approach for efficiently solving larger problem instances. Finally, Section~\ref{sec:conclusion} summarizes the main findings of this work and outlines promising directions for future research.

\section{Related Work}
\label{sec:related}


Learning interpretable classifiers based on Boolean logic has a long tradition in machine learning, including circuit learning, inductive logic programming, rule induction, and associative classification. DNF classifiers have been extensively studied from a theoretical perspective~\cite{kearns1994introduction,valiant1984theory} and are implicitly produced by many practical rule learning systems. 
However, most existing approaches do not explicitly optimize for cognitive simplicity under global optimality guarantees. In particular, inductive logic programming~\cite{
ILP-30,muggleton1994inductive}, associative classification~\cite{cheng2006discriminative,yin2003cpar},
classical rule induction methods~\cite{cohen1995fast,jf:Book-Nada} and modern rule learning algorithms~\cite{ding2022efficient,huynh2023efficient,lakkaraju2016interpretable,wang2017bayesian} typically rely on greedy or sequential covering strategies, which often yield large rule sets without explicit global control of sparsity or misclassification.

More recently, optimization-based approaches have been proposed to explicitly balance predictive accuracy and model simplicity. MILP has emerged as a powerful framework in this context, as it enables the direct encoding of logical constraints in combination with global optimization objectives.
Several works represent trained neural networks as MILPs in order to model piecewise-linear activations and analyze network behavior~\cite{fischetti2018deep}.
Related approaches formulate the training of sparse neural networks directly as MILP problems by encoding binary activations and sparsity constraints while optimizing continuous or quantized weights~\cite{rosenhahn2023optimization}.
Furthermore, rule-based models have been optimized via column generation techniques to handle large candidate spaces~\cite{dash2018boolean,wei2019generalized}.


Only a limited number of optimization-based methods explicitly target sparse DNF classifiers as structured rule networks. Existing approaches often rely on pre-mined candidate rules, explicit pattern enumeration, or decomposed optimization strategies to manage combinatorial complexity~\cite{su2016learning,wang2015learning}.
In contrast, our current formulations adopt a predefined DNF network structure that maps binary inputs directly to conjunctions and optimizes all rules simultaneously within a single MILP.
This provides explicit control over sparsity and misclassification while maintaining exact Boolean semantics and a clear rule-network interpretation.


Recent work has employed SAT-based methods for learning Boolean networks with layered structures in negation normal form (NNF)~\cite{seip2025towards}, which may be viewed as a generalization of DNF. This approach encodes network behavior as propositional satisfiability problems and leverages modern SAT solvers to obtain logically consistent solutions. 
However, while SAT solvers provide strong guarantees of logical consistency,
they do not naturally support optimization objectives such as accuracy--sparsity trade-offs, which are directly addressed in our MILP formulations. Greedy optimization of NNF network structures has also previously been explored in \cite{beck2021beyond,wang20transparent}, but without any guarantees for optimality.

MILP has also been applied to learning optimal decision trees~\cite{bertsimas2017optimal,gunluk2021optimal}, where splits, feature usage, and depth are optimized jointly.
Although such methods provide strong interpretability guarantees, they rely on hierarchical if--then--else structures with ordered and mutually exclusive decision paths.


\section{Sparse Boolean Rule Networks}
\label{sec:sparse_bool_rn}
To ensure the interpretability of the learned model, we formulate our network using a DNF representation.
A DNF is a logical formula
composed of a disjunction of conjunctions of literals. For example, the DNF expression
\begin{equation}
    (f_1 \land f_2 \land f_6) \lor (f_3 \land f_4) \lor (f_5)
    \label{gen_DNF_formula}
\end{equation}
where $f_1, f_2, f_3, f_4, f_5$ and $f_6$ are Boolean input variables, 
evaluates to \textit{true} if at least one of its conjunctions evaluates to \textit{true}.
From an interpretability perspective, each conjunctive node corresponds to a human-readable rule that defines a
sufficient condition for the positive class. Thus, the overall DNF structure can be viewed as a set of rules where satisfying any one of them yields a positive classification.

Note that a DNF formula can also be interpreted as a two-layer network in which the first layer consists of conjunctive nodes (logical AND gates), and the second layer, which is a single output node,
performs a disjunction (logical OR) over the outputs of the first layer. For example, the above DNF expression corresponds to the network shown below in Figure~\ref{fig:nn_example}.

\subsection{Problem Definition}
Given a training dataset consisting of an $n \times m$ binary feature matrix $X$, where $n$ is the number of samples and $m$ the number of features,
and a binary label vector $\mathbf{y}$ of length $n$, our goal is to learn a DNF classifier that not only minimizes misclassification on the training data but also the complexity of the network for overfitting avoidance and increased interpretability.
Analogous to neural networks, we assume a fixed network architecture with a predefined number of conjunctive nodes $\ell$, and then search for the optimal weights
for this fixed architecture.
More precisely, the proposed DNF network consists of:
\begin{itemize}
    \item an input layer with $m$ original and negated features, 
    \item a hidden layer of $\ell$ conjunctive nodes, each representing a rule, and
    \item a single output node performing a disjunction over the conjunction outputs.
\end{itemize}

In principle, the network is fully connected: each input literal is connected to every conjunctive node, and each conjunctive node is connected to the output node. However, each connection is parameterized by a binary weight
indicating whether the corresponding literal is active for the following conjunction or not.
To promote interpretability and prevent overfitting, we aim to learn a sparse DNF network by also minimizing the number of active weights in the model.
If the optimal classifier requires fewer than $\ell$ hidden nodes, the model automatically replicates the simplest rule\,---\,i.e., the rule with the fewest active weights\,---\,thereby allowing duplicate rules to be removed post hoc while preserving a sparse network.
After training, a sample is classified as positive if at least one conjunctive node evaluates to \textit{true} given the input features and the learned binary weights.
Consequently, when predicting labels for unseen data, only connections with active (i.e. positive) weights should be considered.

\begin{example}

\begin{figure}[ht]
\centering
\begin{minipage}[c][0.31\textheight]{0.4\textwidth}
\centering
\begin{subfigure}[t]{\linewidth}
\centering
\[
\displaystyle
(f_1 \land f_2 \land f_6) \lor (f_3 \land f_4) \lor (f_5)
\]
\vspace{-0.5cm}
\caption{DNF-Formula}
\end{subfigure}
\vfill
\renewcommand{\arraystretch}{1.1}
\begin{subfigure}[c]{\linewidth}
    \centering
     \begin{tabular}{c | c c c}
        \toprule
      \multicolumn{1}{c}{\ }  & $w_{i1}$ & $w_{i2}$ & $w_{i3}$\\
        \midrule
       \quad $f_1\quad$ & {1} & 0 & 0  \\
       $f_2$ & {1} & 0 & 0\\
       $f_3$ & 0 & 1 & 0 \\
       $f_4$ & 0 & 1 & 0 \\
       $f_5$ & 0 & 0 & 1 \\
       $f_6$ & {1} & 0 & 0 \\
        \bottomrule
    \end{tabular}
        \caption{Weight matrix $W$}
        \label{fig:ex_values}
\end{subfigure}
\end{minipage}%
\hfill
\begin{minipage}[c]{0.5\textwidth}
\begin{subfigure}[t]{\linewidth}
  \centering
    \begin{tikzpicture}[
      font=\small,
      inputnode/.style={circle, draw=black!40, fill=white, minimum size=5mm, inner sep=0pt},
      hiddennode/.style={circle, draw=blue!40, fill=blue!20, minimum size=5mm, inner sep=0pt},
      hiddennode1/.style={circle, draw=blue, fill=blue!50, minimum size=5mm, inner sep=0pt},
      outputnode/.style={circle, draw=blue!40, fill=blue!20, minimum size=5mm, inner sep=0pt},
      edge/.style={draw=black!55, line width=0.45pt},
      faint/.style={draw=blue!75, line width=0.45pt},
      faint_dashed/.style={draw=blue!75, line width=0.45pt, dashed},
      highlightg/.style={draw=blue, line width=1.6pt},
      highlightb/.style={draw=blue, line width=1.4pt, dashed}
    ]
    
    \node[inputnode] (x1) {};
    \node[inputnode, below=3mm of x1] (x2) {};
    \node[inputnode, below=3mm of x2] (x3) {};
    \node[inputnode, below=3mm of x3] (x4) {};
    \node[inputnode, below=3mm of x4] (x5) {};
    \node[inputnode, below=3mm of x5] (x6) {};
    
    \node[above=1mm of x1] { $x_{ki}$};
    
    \node[hiddennode1, right=17mm of x1] (h1) {};
    \node[hiddennode, below=15mm of h1] (h2) {};
    \node[hiddennode, below=15mm of h2] (h3) {};
    
    \node[above=1mm of h1] { $h_{kj}$};
    
    \node[outputnode, right=12mm of h2] (y) {};
    \node[above=7mm of y] { $\hat{y_k}$};
    
    \node[above=2mm] at ($(x1)!0.55!(h1)$) { $w_{ij}$};
    
    \draw[faint] (x3) -- (h2);
    \draw[faint] (x4) -- (h2);

    \draw[faint] (x5) -- (h3);

    \foreach \hj in {h1,h2,h3}{
      \draw[edge] (\hj) -- (y);
    }
    
    \draw[highlightg] (x1) -- (h1);
    \draw[highlightg] (x2) -- (h1);
    \draw[highlightg] (x6) -- (h1);
    
    \draw[highlightb] (x3) -- (h1);
    \draw[highlightb] (x4) -- (h1);
    \draw[highlightb] (x5) -- (h1);

    \draw[faint_dashed] (x1) -- (h2);
    \draw[faint_dashed] (x2) -- (h2);
    \draw[faint_dashed] (x5) -- (h2);
    \draw[faint_dashed] (x6) -- (h2);

    \draw[faint_dashed] (x1) -- (h3);
    \draw[faint_dashed] (x2) -- (h3);
    \draw[faint_dashed] (x3) -- (h3);
    \draw[faint_dashed] (x4) -- (h3);
    \draw[faint_dashed] (x6) -- (h3);
    
    \end{tikzpicture}
    \caption{Network illustration}
    \label{fig:nn_example}
\end{subfigure}
\end{minipage}
\caption{Illustration of a DNF formula (a), its conversion to a weight matrix $W$ (b), and the resulting network structure (c), consisting of an input layer $X$ into which the $k$-th training example $X_k$ is fed, a hidden layer $H$ which is connected to the inputs with edges corresponding to the weight matrix, and a predictive node $\hat{y}_k$.
Solid lines show active, and dashed lines show inactive weights. 
Bold lines highlight the computation path for the first conjunctive node.}
\label{fig:example}
\end{figure}
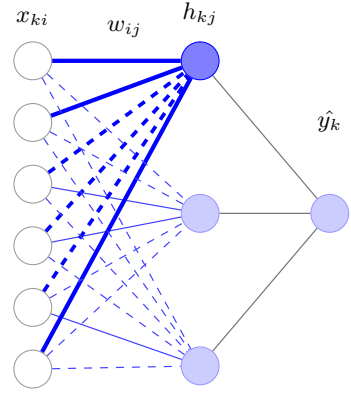
The highlighted connections in Figure~\ref{fig:nn_example} represent the weights of the first conjunctive node, corresponding to the logical rule $(f_1 \land f_2 \land f_6)$, which evaluates to \textit{true} only if all three input features $f_1$, $f_2$, and $f_6$ are equal to 1. 
Using examples $X_1 = [1,0,1,0,0,1]$ and $X_2 = [1,1,0,1,0,1]$ we can observe that the first sample does not satisfy any of these rules, while the second sample fulfills the first. Consequently, due to the disjunctive nature of the DNF structure, the second sample is classified as positive.
\end{example}

Since the overall goal of this work is to learn optimal weights for a sparse DNF network, given an objective function that minimizes both misclassification and the number of active weights for a fixed network structure, the next section presents how this learning problem can be formulated and optimized as a MILP.

\section{Optimizing Sparse Boolean Rule Networks via MILP}
\label{sec:milp_formulation}
To optimize the weights of the sparse DNF network, we propose two MILP formulations.
The first formulation, referred to as the \textit{basic MILP approach}, is formulated directly based on the logical structure shown in Section~\ref{sec:sparse_bool_rn},
while the second formulation, called the \textit{advanced MILP approach}, introduces several improvements to enhance computational efficiency, while ensuring the same level of interpretability.

\subsection{Basic MILP Approach}
The basic idea of the MILP formulation is to explicitly model the value of each hidden node for every training sample, independent of its label, using binary decision variables and linear constraints that capture the logical relationships between all inputs, weights, and hidden nodes.
Therefore, we introduce the following binary decision variables:
\begin{itemize}
    \item $w_{ij} \in \{0,1\}$ 
  encodes whether input  $i$ is included in conjunctive node $j$
    \item $h_{kj} \in \{0,1\}$ 
    represents the output of conjunctive node $j$ for sample $k$
    \item $\hat{y}_{k} \in \{0,1\}$ 
    represents the predicted class for sample $k$
    \item $e_k \in \{0,1\}$ 
    indicates whether sample $k$ is misclassified.
\end{itemize}

\noindent
Using these variables, the MILP problem can be formulated as follows:
\begin{subequations}
\begin{flalign}
    &\quad \textbf{Objective:} \notag \\
    &\quad \min \sum_{k=1}^{n} e_k
    + \left(\alpha \frac{\sqrt{n}}{\sqrt{m}}\right)
    \sum_{i=1}^{m}\sum_{j=1}^{\ell} w_{ij}
    && \label{eq:bas_obj}
\end{flalign}
\begin{alignat}{2}
&\textbf{Subject to:} \notag \\
&\sum_{i=1}^{m}(1 - x_{ki}) w_{ij} \geq 1 - h_{kj}
&\qquad& \forall \quad 1 \leq k \leq n, 1 \leq j \leq \ell
 \label{eq:bas_c1} \\
&\sum_{i=1}^{m}(1 - x_{ki}) w_{ij} \leq m(1 - h_{kj})
&\qquad& \forall \quad 1 \leq k \leq n, 1 \leq j \leq \ell
 \label{eq:bas_c2} \\
&\hat{y}_k \geq h_{kj}
&\qquad& \forall \quad 1 \leq k \leq n, 1 \leq j \leq \ell
 \label{eq:bas_c3} \\
&\hat{y}_k \leq \sum_{j=1}^{\ell} h_{kj}
&\qquad& \forall \quad 1 \leq k \leq n
 \label{eq:bas_c4} \\
&e_k \geq \hat{y}_k - y_k
&\qquad& \forall \quad 1 \leq k \leq n
 \label{eq:bas_c5} \\
&e_k \geq y_k - \hat{y}_k
&\qquad& \forall \quad 1 \leq k \leq n
 \label{eq:bas_c6} \\
&w_{ij}, h_{kj}, \hat{y}_k, e_k \in \{0,1\}
&\qquad& \forall \quad 1 \leq k \leq n, 1 \leq i \leq m, 1 \leq j \leq \ell
 \label{eq:bas_c7}
\end{alignat}
\end{subequations}

\noindent
where the left part of the objective function~\eqref{eq:bas_obj} minimizes misclassifications on the training data, while the right part minimizes the number of active weights in the network to ensure sparsity,
which handles overfitting and further ensures interpretability. The regularization parameter $\alpha$ controls the trade-off between accuracy on the training data and sparsity.
Constraint~\eqref{eq:bas_c1} ensures that rule $j$ predicts \textit{true} for sample $k$ (i.e., $h_{kj} = 1$) if all input literals included in the conjunction (those for which $w_{ij} = 1$) are also \textit{true}.
Conversely, constraint~\eqref{eq:bas_c2} guarantees that if at least one input literal $x_{ki}$ included in the conjunction ($w_{ij} = 1$) is \textit{false} for sample $k$, then rule $j$ is not satisfied for sample $k$.
Constraint~\eqref{eq:bas_c3} ensures that the predicted label $\hat{y}_k$ is \textit{true} if at least one rule $j$ fires for sample $k$ ($h_{kj} = 1$), whereas
constraint~\eqref{eq:bas_c4} guarantees that if no rule $j$ 
is fulfilled for sample $k$ ($\forall j: h_{kj} = 0$), then the predicted class $\hat{y}_k$ must be \textit{false}.
Constraints~\eqref{eq:bas_c5} and~\eqref{eq:bas_c6} define the misclassification variable $e_k$ based on the predicted class $\hat{y}_k$ and the true label $y_k$ for each sample $k$.
Finally, constraint~\eqref{eq:bas_c7} enforces the binary nature of all decision variables in the formulation.

\begin{figure}[t]
\begin{subfigure}[t]{0.45\textwidth}
  \centering
    \begin{tikzpicture}[
      font=\small,
      inputnode/.style={circle, draw=black!40, fill=white, minimum size=5mm, inner sep=0pt},
      hiddennode/.style={circle, draw=blue!40, fill=blue!20, minimum size=5mm, inner sep=0pt},
      outputnode/.style={circle, draw=blue!50, fill=blue!20, minimum size=5mm, inner sep=0pt},
      edge/.style={draw=black!55, line width=0.45pt},
      faint/.style={draw=blue!45, line width=0.35pt},
      highlightg/.style={draw=green!70!black, line width=1.6pt},
      highlightb/.style={draw=black, line width=1.4pt}
    ]
    
    \node[inputnode] (x1) {1};
    \node[inputnode, below=3mm of x1] (x2) {0};
    \node[inputnode, below=3mm of x2] (x3) {1};
    \node[inputnode, below=3mm of x3] (x4) {0};
    \node[inputnode, below=3mm of x4] (x5) {0};
    \node[inputnode, below=3mm of x5] (x6) {1};
    
    \node[above=1mm of x1] { ${x_{1i}}$};
    
    \node[hiddennode, right=15mm of x1] (h1) {};
    \node[hiddennode, below=15mm of h1] (h2) {};
    \node[hiddennode, below=15mm of h2] (h3) {};
    
    \node[above=1mm of h1] { ${h_{1j}}$};
    
    \node[outputnode, right=12mm of h2] (y) {};
    \node[above=7mm of y] { $\hat{y_1}$};
    
    \node[above=2mm] at ($(x1)!0.55!(h1)$) { $w_{ij}$};
    
    \foreach \xi in {x1,x2,x3,x4,x5,x6}{
      \foreach \hj in {h1,h2,h3}{
        \draw[faint] (\xi) -- (\hj);
      }
    }
    
    \foreach \hj in {h1,h2,h3}{
      \draw[edge] (\hj) -- (y);
    }
    
    
    \end{tikzpicture}
          \caption{Basic approach}
  \label{fig:nn_bas_example}
\end{subfigure}
\hfill
\begin{subfigure}[t]{0.45\textwidth}
  \centering
    \begin{tikzpicture}[
      font=\small,
      inputnode/.style={circle, draw=black!40, fill=white, minimum size=5mm, inner sep=0pt},
      hiddennode/.style={circle, draw=blue!40, fill=blue!20, minimum size=5mm, inner sep=0pt},
      outputnode/.style={circle, draw=blue!50, fill=blue!20, minimum size=5mm, inner sep=0pt},
      edge/.style={draw=black!55, line width=0.45pt},
      faint/.style={draw=blue!45, line width=0.35pt},
      highlightg/.style={draw=green!70!black, line width=1.6pt},
      highlightb/.style={draw=black, line width=1.4pt}
    ]
    
    \node[inputnode] (x1) {1};
    \node[inputnode, below=3mm of x1] (x2) {0};
    \node[inputnode, below=3mm of x2] (x3) {1};
    \node[inputnode, below=3mm of x3] (x4) {0};
    \node[inputnode, below=3mm of x4] (x5) {0};
    \node[inputnode, below=3mm of x5] (x6) {1};
    
    \node[above=1mm of x1] { ${x_{1i}}$};
    
    \node[hiddennode, right=15mm of x1] (h1) {};
    \node[hiddennode, below=15mm of h1] (h2) {};
    \node[hiddennode, below=15mm of h2] (h3) {};
    
    \node[above=1mm of h1] { ${h_{1j}}$};
    
    \node[outputnode, right=12mm of h2] (y) {};
    \node[above=7mm of y] { $\hat{y_1}$};
    
    \node[above=2mm] at ($(x1)!0.55!(h1)$) { $w_{ij}$};
    
    \foreach \xi in {x2,x4,x5}{
      \foreach \hj in {h1,h2,h3}{
        \draw[faint] (\xi) -- (\hj);
      }
    }
    
    \foreach \hj in {h1,h2,h3}{
      \draw[edge] (\hj) -- (y);
    }

    
    \end{tikzpicture}
    
    \caption{Advanced approach}
  \label{fig:nn_adv_example}
\end{subfigure}
\caption{DNF network structure for basic and advanced constraint formulation.}
\end{figure}
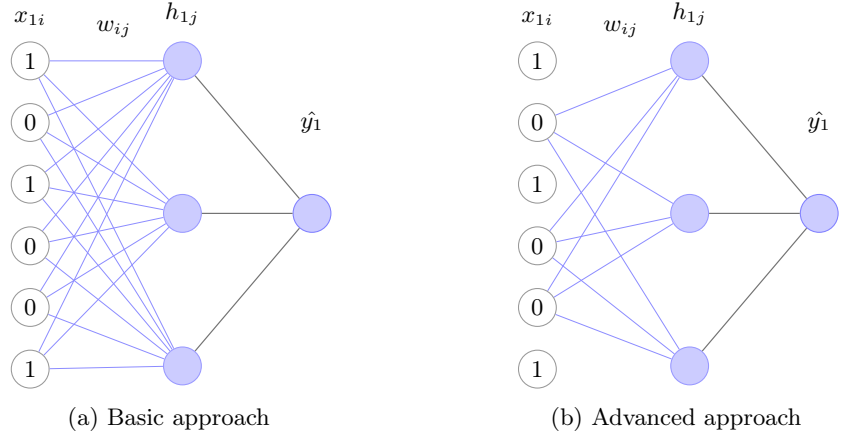


\subsection{Advanced MILP Approach}

As illustrated in Figure~\ref{fig:nn_bas_example}, the basic MILP approach requires adding constraints for every combination of input feature, conjunctive node, and training sample,
regardless of the input feature values, resulting in a fully connected network. Moreover, constraints must be added to restrict the hidden node values from both sides, since samples with positive ($y_k = 1$) and negative class ($y_k = 0$) labels, hereafter referred to as positive and negative samples, are treated identically. More precisely, for the decision variables representing hidden node activations, whose values depend on both the input features and the weights, two constraints are always introduced: one ensuring that the variable is set to one whenever the combination of input features and weights activates the node, and another ensuring that it is set to zero otherwise. Consequently, the number of constraints grows rapidly with the size of the instance, leading to substantial scalability challenges.
The resulting large number of binary decision variables may lead to large branch-and-bound trees during optimization, making this basic formulation computationally expensive, in particular for larger datasets. To be more precise, using this formulation we need to optimize over $m \times \ell + n \times (\ell+2)$ binary decision variables.

To address these limitations of the basic MILP approach, we propose an advanced MILP approach, which introduces the following improvements:
\begin{itemize}
    \item Removal of decision variables for predicted class of all samples and hidden nodes of negative samples
    \item Replacement of binary decision variables for the remaining hidden nodes and misclassification indicators with continuous variables
    \item Reformulation of constraints to ensure that the continuous decision variables attain binary values in all feasible solutions
    \item Formulation of only the constraints that are necessary based on the label and feature values of each sample
\end{itemize}
Using these improvements, the advanced MILP model is formulated as follows:
\begin{subequations}
    \begin{flalign}
    &\quad \textbf{Objective:} \notag \\
    &\quad \min \sum_{k=1}^{n} e_k
    + \left(\alpha \frac{\sqrt{n}}{\sqrt{m}}\right)
    \sum_{i=1}^{m}\sum_{j=1}^{\ell} w_{ij}
    && \label{eq:adv_obj}
    \end{flalign}
    \begin{alignat}{2}
        &\textbf{Subject to:} \notag \\
        &w_{ij} \leq 1 - h_{kj}
        &\quad& \forall\ 1 \leq k \leq n, 1 \leq i \leq m, 1 \leq j \leq \ell:\ y_k=1,\ x_{ki}=0
        \label{eq:adv_c1} \\
        &e_k \geq 1 - \sum_{j=1}^{\ell} h_{kj}
        &\quad& \forall\ 1 \leq k \leq n:\ y_k=1
        \label{eq:adv_c2} \\
        &e_{k} \geq 1 - \sum_{i=1: x_{ki} = 0}^{m} w_{ij} 
        & \quad& \forall\ 1 \leq k \leq n, 1 \leq j \leq \ell:\ y_k=0 \label{eq:adv_c3} \\
        &w_{ij} \in \{0,1\}
        &\quad& \forall\ 1 \leq i \leq m, 1 \leq j \leq \ell
        \label{eq:adv_c5} \\
        &h_{kj}, e_k \geq 0
        &\quad& \forall\ 1 \leq k \leq n, 1 \leq j \leq \ell
        \label{eq:adv_c6}
    \end{alignat}
\end{subequations}

\noindent
While the objective function~\eqref{eq:adv_obj} remains unchanged compared to the basic formulation,
the constraints have been significantly modified to improve computational efficiency.
When constructing the constraints, we distinguish between samples based on their class labels.
Constraint~\eqref{eq:adv_c1} ensures that, for positive samples, rule $j$ is not fulfilled ($h_{kj} = 0$) if sample $k$ contains a \textit{false} input literal ($x_{ki}=0$) that is included in the conjunctive node ($w_{ij}=1$).

As the objective function automatically pushes all $h_{kj}$ variables for positive samples to be fulfilled in order to minimize misclassifications, we only need to restrict the value of $h_{kj}$ from above. Furthermore, we observe that only weight connections to negative input features must be considered, as positive input features cannot force hidden nodes to be unfulfilled, regardless of the weight values. By exploiting this separation, we avoid introducing this constraint for every combination of hidden nodes and decision variables for negative samples, as well as constraints involving weight connections to positive features for positive samples.
Constraint~\eqref{eq:adv_c2} ensures that, for positive samples ($y_k=1$), the misclassification variable $e_k$ is set to \textit{true} if no rule $j$ is fulfilled by sample $k$ ($\forall j: h_{kj} = 0$). Again, we only need to restrict the value of the misclassification decision variable from below, because the objective function automatically pushes it to be zero.
For negative samples we can reformulate the constraints so that we only need the weight variables to define whether a sample is misclassified, thereby removing all decision variables for the hidden nodes of negative samples.
This is possible because the values of the remaining decision variables for negative samples depend only on the values of the weights. More precisely, the reformulated constraint~\eqref{eq:adv_c3} directly ensures that a negative sample is misclassified, if there exists at least one hypothetical node with no active weight connection from a negative input feature. Interestingly, no additional decision variables for the hidden nodes are required, as the misclassification can be mapped directly through the weights. Furthermore, we again only need to restrict the misclassification error from one side, because the objective function automatically pushes it to be zero. With these reformulations, we not only eliminate all decision variables for the hidden nodes of the negative samples, but also only need to consider connections from negative input features.
Finally, constraints~\eqref{eq:adv_c5} and~\eqref{eq:adv_c6} define the binary nature of the weight variables and the continuous nature of the hidden node and misclassification variables, respectively.
With these reformulations, all continuous decision variables are forced to binary values in any feasible solution.

Figure~\ref{fig:nn_adv_example} illustrates that the advanced MILP approach only requires constraints for negative input features and distinguishes between positive and negative samples when constructing the constraints for the hidden nodes and misclassification variables. This substantially reduces the number of constraints in the model. In addition, the reformulation using continuous decision variables, which reduces the number of binary decision variables to $m \times \ell$ and the total number of decision variables to $m \times \ell + n^\oplus \times \ell + n$, where $n^\oplus$ represents the number of positive samples, leads to smaller branch-and-bound trees during optimization, thereby making the advanced formulation computationally more efficient, particularly for larger datasets.

\subsection{Aggregated Constraints}
\label{subsec:aggr_con}
An aggregated constraint combines multiple individual constraints into one more compact aggregated formulation, thereby reducing the overall number of constraints.
Although \textit{linear programming} (LP) theory generally suggests that aggregated constraints lead to weaker LP relaxations at branch and bound nodes, partial aggregation can still be beneficial in practice, as it reduces the total number of constraints required to ensure model correctness~\cite{wolsey2020integer}.
More specifically, we aim to investigate the trade-off between a more compact model formulation with fewer constraints and the potentially weaker relaxation induced by the aggregation.
This trade-off is particularly relevant in combination with the heuristic and branching techniques employed by modern MILP solvers such as Gurobi,\footnote{\url{https://www.gurobi.com}} which is used as the optimization solver in our computational evaluation.

More precisely, constraints~\eqref{eq:adv_c1} can be replaced by the following aggregated formulation:
\begin{equation}
     \sum_{i=1: x_{ki} = 0}^{m} \frac{w_{ij}}{n^\ominus_k}
    + h_{kj} \leq 1 \quad \forall\ 1 \leq k \leq n, 1 \leq j \leq \ell:\ y_k=1,
    \label{eq:aggr_con}
\end{equation}
where 
$n^\ominus_k$ represents the number of negative input features of sample $k$.
When using constraints~\eqref{eq:aggr_con}, the hidden-node decision variables must be explicitly restricted to binary values in order to ensure model correctness, since the aggregated formulation no longer guarantees integrality implicitly through the constraint structure.
Nevertheless, because Gurobi employs advanced branching, cutting-plane, and primal heuristic strategies, the aggregated formulation may still provide competitive computational performance despite the weaker LP relaxation.

\section{Experimental Evaluation}
\label{sec:exp_eval}
In this section, we first compare the basic and advanced MILP approaches regarding their computational efficiency and solution quality on artificial datasets, to analyze the efficiency gain of the advanced formulation using different time limits for the solver.
Secondly, we evaluate the performance of the advanced MILP approach against state-of-the-art interpretable classification methods on real-world UCI datasets.
In particular, we compare the accuracy of our approach against the SAT-based method of \cite{seip2025towards}, where the authors also use a DNF structure for classification,
the \alg{Lord} method \cite{huynh2023efficient}, which learns locally optimal rules and outperforms other rule-based methods, as well as the well-known rule learning algorithm
\alg{Ripper} \cite{cohen1995fast} and the decision tree learner \alg{Cart} \cite{breiman2017classification}.
We computed the accuracy of our MILP approaches and all compared algorithms using 10-fold cross-validation.

\subsection{Experimental Setup}
We implemented both MILP formulations in Python, using the Gurobi 12.0.0 MILP solver to optimize the MILP models shown in Section~\ref{sec:milp_formulation}. 
We let our experiments run on a single core of an AMD EPYC 7543 processor running at 3.9 GHz.
To compare and evaluate the proposed MILP approaches against state-of-the-art interpretable classification methods, we selected the following datasets.
\begin{itemize}
    \item \textbf{Artificial instance set:} This instance set consists of 20 datasets with 10 binary features and 1024 samples each.
    These datasets were artificially generated by \cite{beck2021empirical} and have also been used in \cite{beck2021beyond,beck2021structuring,seip2025towards}.
    We use the same number of hidden nodes as used in \cite{beck2021empirical} and \cite{seip2025towards}, which is 20 hidden nodes, to ensure a fair comparison.
    We identify these datasets by ordering them based on the random seed used during their generation and referring to them as d01 to d20. We selected these datasets to compare our approaches under different time limits, as all instances have the same size and can therefore be evaluated using identical time limits across all experiments.
    \item \textbf{UCI Datasets:} To evaluate our advanced approach on real-world data, we selected nine widely used benchmark datasets from the UCI repository~\footnote{https://archive.ics.uci.edu}.
    The number of binarized features ranges from 17 to 116, while the number of samples ranges from 435 to 8124.
    All datasets represent binary classification problems, except for the car-evaluation dataset, which we transformed into a binary task by predicting whether a sample belongs to the most frequent class.
    For our approach, all input attributes were binarized by first discretizing them with \alg{Fusinter} \cite{zighed1998fusinter} and then using a one-hot encoding, as described in \cite{huynh2023efficient}. The resulting binary datasets were used in all experiments.
    Additional experiments on the original datasets with rule- and tree-based learning algorithms yielded similar results, so we used the binarized datasets for all methods to ensure a fair comparison.
\end{itemize}

\subsection{Parameter Tuning}
As we use a fixed network architecture with a predefined number of hidden nodes, we only need to tune the sparsity parameter $\alpha$ for both MILP approaches. The reason for using a fixed network architecture is that the long-term goal of this work is to apply the optimization approach to individual components of a decomposed deep rule network, where the structure is predefined. We performed the parameter tuning on a training dataset where we included five artificial datasets and one UCI dataset.
We used the automatic parameter tuning tool \alg{Irace}~\cite{lopez2016irace}, which uses an iterated racing procedure to identify the best value for $\alpha$.
We set the experimental budget to 1000 algorithm runs while using a time limit of 1200 seconds for solving each MILP, where we used the advanced approach.
We defined the candidate values for $\alpha$ to be in the range of $[0.001, 0.5]$.
The best identified value for $\alpha$ was $0.06$, which we used for all MILP approaches in all experiments.

\subsection{Comparison of Basic and Advanced MILP Approaches}
To show the improvements found with the advanced MILP approach and to analyze the impact of the aggregation of constraints, we first tested our approaches on the artificial datasets using the same time limits and parameter configuration for each version.
We therefore let both variants run using 10-fold cross-validation as described above and reported the results in Table~\ref{tab:comp_art}.

\begin{table}[ht]
\centering
\caption{Comparing accuracies of basic and advanced MILP approaches on artificial datasets.
The best accuracy per instance is marked in \textbf{bold}, while the best accuracy for the shorter time limit is in \textit{italic}.}
{\setlength{\tabcolsep}{7pt} 
\begin{tabular}{c|rrr|rrr}
\toprule
        \multirow[b]{2}{*}{Dataset}
    & \multicolumn{3}{c}{120 s} & \multicolumn{3}{c}{360 s} \\
   \cmidrule{2-4}\cmidrule{5-7}
    & basic & advanced & aggregated & basic & advanced & aggregated\\
\midrule
    d01 & 96.00 & 96.49 & \textit{98.05} & 99.90 & \textbf{100.00} & \textbf{100.00} \\ 
    d02 & 84.66 & 97.85 & \textit{98.05} & 97.55 & 99.12 & \textbf{99.22} \\ 
    d03 & 96.49 & \textit{\textbf{100.00}} & \textit{\textbf{100.00}} & \textbf{100.00} & \textbf{100.00} & \textbf{100.00} \\ 
    d04 & 84.76 & 89.94 & \textit{94.44} & 99.02 & 99.71 & \textbf{99.71} \\ 
    d05 & 88.18 & 91.21 & \textit{94.44} & 94.54 & 98.64 & \textbf{98.83} \\ 
    d06 & 83.50 & 91.79 & \textit{97.94} & 94.94 & \textbf{99.71} & \textbf{99.71} \\ 
    d07 & 96.68 & 98.05 & \textit{98.34} & 99.22 & \textbf{100.00} & 99.90 \\ 
    d08 & 94.64 & 99.61 & \textit{99.90} & 99.61 & 99.90 & \textbf{99.90} \\ 
    d09 & 80.96 & 93.46 & \textit{97.66} & 96.49 & 99.41 & \textbf{99.51} \\ 
    d10 & 96.31 & \textit{\textbf{100.00}} & 99.80 & \textbf{100.00} & \textbf{100.00} & \textbf{100.00} \\ 
    d11 & 91.12 & 88.50 & \textit{97.46} & 98.54 & 99.12 & \textbf{99.51} \\ 
    d12 & 89.25 & 96.49 & \textit{98.53} & 97.16 & 98.73 & \textbf{98.93} \\ 
    d13 & 90.14 & 89.16 & \textit{96.88} & 96.68 & 99.32 & \textbf{99.51} \\ 
    d14 & 89.54 & 88.39 & \textit{94.44} & 98.44 & 97.95 & \textbf{98.83} \\ 
    d15 & 98.44 & \textit{\textbf{100.00}} & \textit{\textbf{100.00}} & \textbf{100.00} & \textbf{100.00} & \textbf{100.00} \\ 
    d16 & 97.66 & \textit{\textbf{100.00}} & \textit{\textbf{100.00}} & \textbf{100.00} & \textbf{100.00} & \textbf{100.00} \\ 
    d17 & 85.56 & 87.91 & \textit{97.65} & 99.32 & 99.61 & \textbf{99.71} \\ 
    d18 & 89.95 & 92.18 & \textit{96.09} & 97.56 & 99.22 & \textbf{99.61} \\ 
    d19 & 88.29 & \textit{96.19} & 96.09 & 96.39 & 98.44 & \textbf{99.22} \\ 
    d20 & 89.68 & 93.56 & \textit{98.54} & \textbf{99.61} & 99.32 & \textbf{99.61} \\ 
\midrule
    \textbf{$\emptyset$ rank}  & 5.85 & 4.25 & \textit{3.30} & 2.85 & 1.75 & \textbf{1.05}\\
    \textbf{$\emptyset$ acc} & 90.59 & 94.54 & \textit{97.72} & 98.25 & 99.41 & \textbf{99.59}  \\
\bottomrule
\end{tabular}
}
\label{tab:comp_art}
\end{table}


In Table~\ref{tab:comp_art}, the first column lists the dataset identifiers.
The second, third, and fourth columns report the accuracies achieved by the basic MILP approach, the advanced MILP approach, and the advanced approach with aggregated constraints, respectively, under a time limit of 120 seconds.
The last three columns present the corresponding results for a time limit of 360 seconds.

As can be observed, both variants of the advanced approach consistently outperform the basic MILP formulation under both time limits.
When comparing the average accuracies, the performance gap between the advanced and the basic formulations increases as the time limit decreases.
This observation further supports the claim that the advanced MILP formulation is computationally more efficient than the basic model.

Moreover, the results indicate that, particularly under tight time limits, the trade-off of using a model with fewer constraints outweighs the benefits of the stronger formulations obtained from individual constraints.
Overall, the best results are achieved by the advanced approach using aggregated constraints with a time limit of 360 seconds.
In summary, the advanced approach clearly outperforms the basic one for both considered time limits, while the variant with aggregated constraints even further improves efficiency as it reduces the number of constraints.
Although the performance difference between the aggregated and non-aggregated versions decreases as the time limit increases, which is consistent with our expectations, we decided to use the aggregated-constraint formulation for the comparison with well-known algorithms on real-world datasets.
The reason is that, especially for larger datasets, computational efficiency becomes increasingly important.

\subsection{Comparing Advanced MILP Approach to Existing Approaches}

As previously mentioned, we compare the accuracy of our method to  the SAT-based method for learning DNF structures from~\cite{seip2025towards}, the \alg{Lord} rule discoverer by~\cite{huynh2023efficient}
and the classic algorithms \alg{Ripper}~\cite{cohen1995fast} and \alg{Cart}~\cite{breiman2017classification}.
%
These algorithms
have achieved highly competitive results on the tested datasets and can therefore serve as well-established reference points for our approach.

\begin{table}[t]
\centering
\caption{Comparing accuracies of the advanced MILP approach with aggregated constraints to existing algorithms on UCI datasets. The best accuracy per instance is marked in \textbf{bold}.}
{\setlength{\tabcolsep}{7pt} 
\begin{tabular}{lr|rrrrr}
    \toprule
    \multirow[b]{2}{*}{Dataset}
    &     \multirow[b]{2}{*}{\makecell{Time\\limit}} &\multicolumn{4}{c}{Approaches}\\
 \cmidrule{3-7}
  & & \alg{Milp}& \alg{Sat} & \alg{Lord} & \alg{Ripper} & \alg{Cart} \\ 
    \midrule
    breast & 300 & \textbf{95.99} & 94.56 & 94.71 & 94.28 & 93.42 \\ 
    car-evaluation & 1200 & \textbf{99.36} & 98.67 & 98.20 & 96.70 & 98.03 \\ 
    kr-vs-kp & 7200 & \textbf{99.66} & 99.50 & 99.50 & 99.37 & 99.44 \\ 
    monk-1 & 30 &\textbf{100.00} & \textbf{100.00} & \textbf{100.00} & 93.70 & \textbf{100.00} \\ 
    monk-2 & 180 & \textbf{100.00} & 98.67 & 87.20 & 73.54 & 97.84 \\ 
    monk-3 & 1800 & \textbf{98.92} & 96.20 & 97.65 & 98.56 & 98.92 \\ 
    mushroom & 600 & \textbf{100.00} & \textbf{100.00} & \textbf{100.00} & 99.99 & 99.28 \\ 
    tic-tac-toe & 600 & \textbf{100.00} & 99.27 & 98.22 & 97.91 & 93.22 \\ 
    vote & 200 & \textbf{96.56} & 91.03 & 95.41 & 94.94 & 95.63 \\ 
    \midrule
    \textbf{$\emptyset$ rank}  & & \textbf{1.00} & 2.56 & 2.67 & 4.33 & 3.44\\
    \textbf{$\emptyset$ acc}  & & \textbf{98.97} & 97.54 & 96.77 & 94.33 & 97.31  \\
    \bottomrule
\end{tabular}
}

\label{tab:comp_uci_other}
\end{table}
Table~\ref{tab:comp_uci_other} compares the performance of the advanced MILP approach to its competitors on the real-world UCI datasets.
As indicated by the bold values, our model matches or surpasses the predictive accuracy of all 
benchmark algorithms on all tested datasets.
This further strengthens the argument for the robustness of our method, even on real-world instances.
More precisely, the advanced MILP approach achieves a mean accuracy of nearly 99 \% on the UCI datasets, outperforming all compared algorithms by more than 1 \% on average.
Since our method is based on an exact optimization framework, it naturally cannot compete with the solution times of heuristic approaches
such as rule learners and decision tree learners which, however, was not the goal of this work. 

\section{Informed Initialization and Variable Reduction}
\label{sec:matheuristic}
The main objective of this combined approach is to efficiently tackle larger problems. This section is divided into two parts. First, we investigate the impact of warm-starting the MILP optimization with an initial solution that has been obtained by a quick greedy optimization, in our case the solution obtained by \alg{Ripper}. Second, we analyze the effect of variable reduction by dropping unused weight connections from the \alg{Ripper} solution. 
To evaluate the effectiveness of the proposed approach, we test the advanced MILP formulation with aggregated constraints, both with and without the proposed enhancements, on the UCI dataset \texttt{adult}, which consists of more than 48,000 samples and 130 binarized features. All experiments are conducted using 10-fold cross-validation.

\subsection{\alg{Ripper}-Based Initialization}
Table~\ref{tab:init} shows the impact of warm-starting the MILP optimization with the \alg{Ripper} solution. As we can see, using the initialization leads to a significant improvement on the test accuracy, with an even larger difference at shorter time limits. While the solution obtained from random initializations slowly improves over time and can be expected to reach the same level at higher time-outs, an informed initialization helps to jump-start the optimization algorithm at a much higher level.
However, the warm-started MILP optimization also constantly improves the initial \alg{Ripper} solution, already by more than 0.3 \% using a time limit of 300 seconds, highlighting the efficiency of the advanced MILP formulation in refining existing solutions. 

\begin{table}[ht]
\caption{Accuracies on \texttt{adult} dataset with and without \alg{Ripper}-Initialization.}
\label{tab:init}
\centering
{\setlength{\tabcolsep}{7pt} 
\begin{tabular}{c|rrrrrr}
  \toprule
    \multirow[b]{2}{*}{\makecell{\alg{Ripper} \\Initialization}}
     &\multicolumn{6}{c}{Time Limit}\\
 \cmidrule{2-7}
 & 300 & 600 & 900 & 1800 & 3600 & 7200 \\ 
  \hline
    No & 76.07 & 76.07 & 76.07 & 76.63 & 78.63 & 80.28 \\
    Yes & 84.75 & 84.77 & 84.87 & 84.87 & 84.96 & 85.17 \\
\bottomrule
\end{tabular}
}
\end{table}

\subsection{\alg{Ripper}-Based Variable Reduction}
\begin{figure}[htbp]
    \centering
    \includegraphics[width=0.8\textwidth]{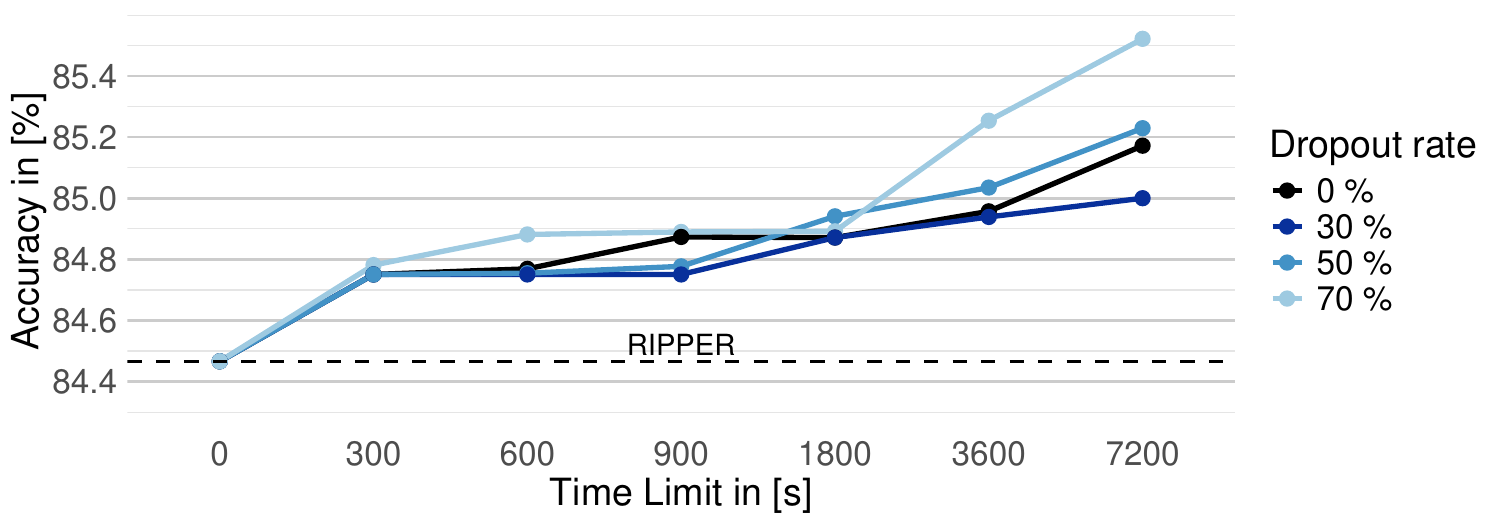}
    \caption{Accuracy on \texttt{adult} dataset by time limit and dropout rate.}
    \label{fig:accuracy_plot}
\end{figure}

Figure~\ref{fig:accuracy_plot} visualizes the impact of different dropout rates for the unused weight connections from the \alg{Ripper} solution, on the advanced MILP formulation using aggregated constraints and the \alg{Ripper}-Initialization. 
As expected, for very small time limits, configurations with higher dropout rates perform better because they reduce the solution space more aggressively. Interestingly, even for a higher time limit of 7200 seconds, the best result is achieved
using a dropout rate of 70 \%, thereby improving the original \alg{Ripper} solution by more than 1 \%, achieving a final test accuracy of 85.52 \%. Similar results on dropout rates of 60 and 80 \% strengthen the argument that, for our problem, it is useful to break symmetries in the weight matrix, since, without such variable reduction, each rule may be represented by any hidden node.

Overall, these findings indicate that it is not only beneficial to warm-start the MILP solver with a good initial solution, but that it is also useful to use variable reduction in order to decrease the solution space. 
Consequently, similar decomposition and variable reduction strategies represent a promising direction for future research.

\section{Conclusion and Future Work}
\label{sec:conclusion}
In summary, we demonstrated that the advanced MILP approach clearly outperforms the basic formulation.
The main reason for this improvement lies in the reformulation of the constraints and the removal of unnecessary decision variables. Furthermore, we showed that the use of aggregated constraints can further improve the computational efficiency of the model, thereby enhancing the overall performance of the approach.
Additionally, our advanced MILP approach consistently outperforms existing algorithms in terms of accuracy on real-world UCI datasets.
As discussed in Section~\ref{sec:sparse_bool_rn}, the major advantage of MILP-based models is that sparsity can be directly incorporated into the objective function.
This not only mitigates overfitting but also enhances interpretability by encouraging simpler, shorter rules within the optimization process.
Through the reformulations introduced in the advanced approach, we have also improved the model’s computational efficiency for modern MILP solvers such as Gurobi.
In addition, we showed that, particularly under tight time limits and for larger datasets, initializing the advanced MILP formulation with a greedily obtained solution and pruning part of the unused weight connections improves efficiency by mitigating network symmetries.

Even though these techniques already allow us to scale up to large datasets such as \texttt{adult}, scalability is still the main limitation of our current approach.
A particularly promising direction is the application of Benders decomposition, which would allow us to divide the overall problem into a master problem and a set of subproblems (one per sample).
We intend to explore this idea further in future work, with the goal of extending it towards  deeper Boolean rule network architectures, with several consecutive AND and OR layers \cite{beck2021empirical}.
%
To this end,
we aim to iteratively optimize selected parts of the network using decomposition strategies similar to those developed in previous work by the main author~\cite{pammer2026matheuristic}.


\section*{Acknowledgments}
This research was funded by the Austrian Science Fund (FWF) 10.55776/COE12.

\bibliographystyle{splncs04}
\bibliography{references}
%




\end{document}